\documentclass[11pt]{article}

\PassOptionsToPackage{table}{xcolor}
\usepackage[preprint]{acl}
\usepackage{times}
\usepackage{latexsym}
\usepackage[T1]{fontenc}
\usepackage[utf8]{inputenc}
\usepackage{microtype}
\usepackage{inconsolata}
\usepackage{graphicx}

\usepackage{amsmath,amssymb}
\usepackage{booktabs}
\usepackage{stfloats}
\usepackage{multirow}
\usepackage{xcolor}
\usepackage{tikz}
\usetikzlibrary{arrows.meta,positioning,fit,backgrounds,calc}
\usepackage{pgfplots}
\pgfplotsset{compat=1.17}
\usepgfplotslibrary{groupplots}

\newcommand{\ours}{\textsc{FedVP}}

\definecolor{oursrow}{RGB}{232,240,250}
\definecolor{cA}{RGB}{31,119,180}
\definecolor{cB}{RGB}{214,39,40}
\definecolor{cC}{RGB}{44,160,44}
\definecolor{cD}{RGB}{127,127,127}

\title{FedVP: Federated Learning across Heterogeneous Models\\by Pooling Locally Trained Verifiers}

\author{%
  \begin{tabular}{@{}c@{\hspace{2em}}c@{}}
    Yifan Li$^{1}$ & Xi Chen$^{1,2}$\\
    {\small\normalfont\texttt{yl3548@bath.ac.uk}} &
    {\small\normalfont\texttt{xc841@bath.ac.uk/x\_chen@fudan.edu.cn}}
  \end{tabular}\\[0.4em]
  $^{1}$University of Bath\\
  $^{2}$Artificial Intelligence Innovation and Incubation Institute,\\
  Fudan University}
\begin{document}
\maketitle

% ---------------------------------------------------------------------------
\begin{abstract}
Conventional federated learning relies on the shared model backbone to aggregate client updates, limiting collaboration among institutions that deploy large language models (LLMs) from different families and at different scales.

We propose \textbf{Federated Verifier Pool} (\ours{}), a new federated learning paradigm that enables heterogeneous clients to collaborate through shared verification rather than parameter aggregation. Each client trains a verifier using local data and responses generated by its chosen LLM backbone. The server pools these independently trained verifiers and combines their judgments to select answers without merging model parameters. This allows both LLM backbones and verifiers to differ in architecture and size, while keeping the LLM backbones frozen and deriving collaborative gains from improved answer selection.

We evaluate \ours{} across five public medical question-answering datasets treated as federated clients, using Qwen and LLaMA backbones ranging from 3B to 14B parameters. \ours{} consistently improves performance across all five clients and outperforms the strongest conventional federated baselines by 1.46--5.79 percentage points in mean accuracy. Code is available at \url{https://github.com/lyf7374/FedVP}.
\end{abstract}

% ---------------------------------------------------------------------------
\section{Introduction}

\begin{figure*}[t]
\centering
\includegraphics[width=\textwidth]{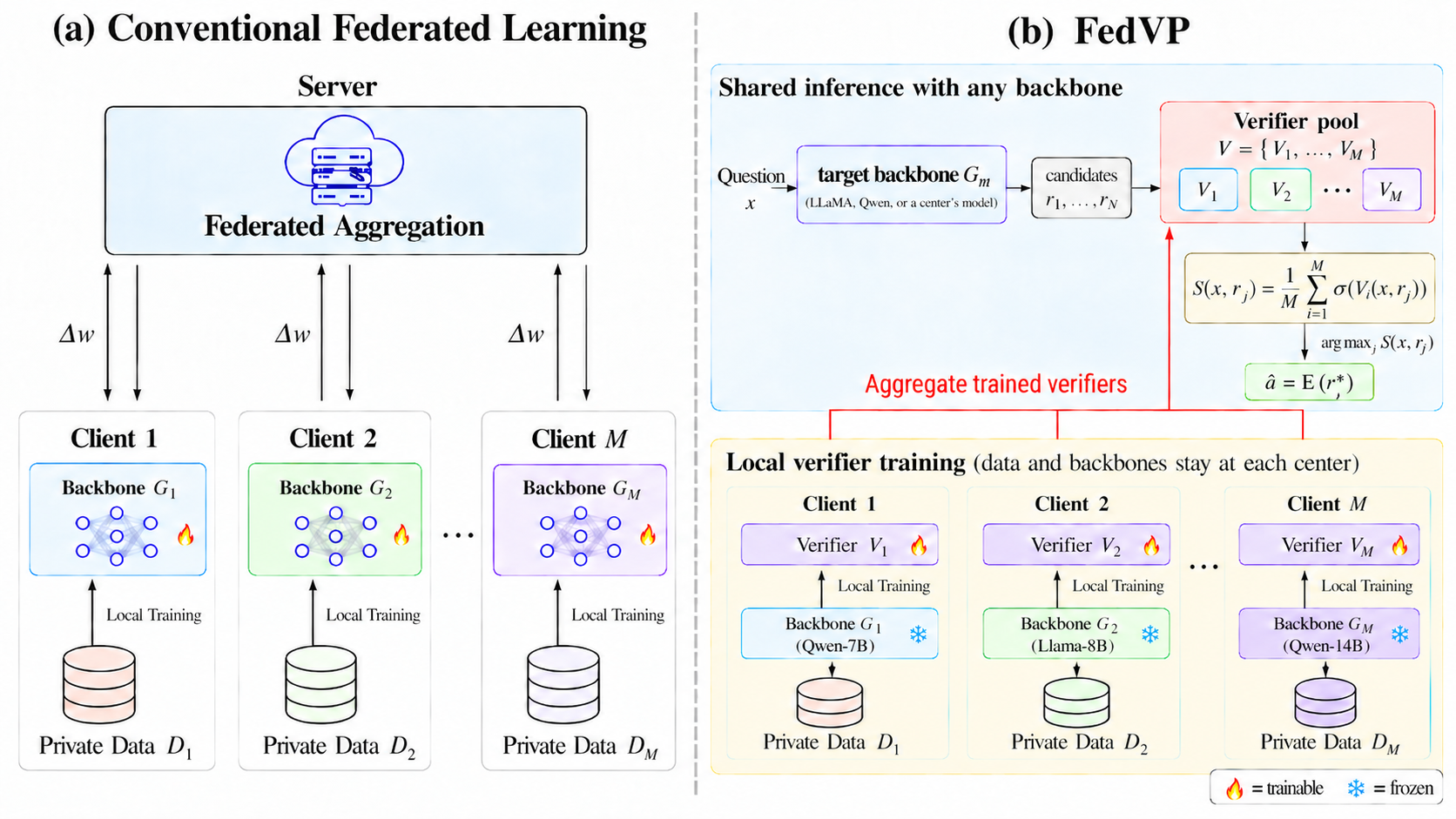}
\caption{Conventional federated fine-tuning versus \ours{}. (a) Conventional FL aggregates updates from clients sharing a common generator backbone. (b) \ours{} enables clients with heterogeneous generators and verifiers to collaborate through a shared pool of independently trained verifiers. The pool collectively evaluates candidate responses from a target generator to select the final answer. The illustrated clients demonstrate the workflow; the evaluated five-center configuration is detailed in Table~\ref{tab:client_models}.}
\label{fig:overview}
\end{figure*}
Federated learning (FL) enables collaborative training across institutions without centralizing their local data \citep{mcmahan2017fedavg,kairouz2021advances}. Recent advances have extended FL to large language models (LLMs) through parameter-efficient fine-tuning, including federated adapter and prompt tuning \citep{zhang2023fedit,sun2024ffalora,wang2024flora,zhao2022fedprompt}. However, participating institutions may differ not only in their local data but also in their choice of LLM backbones, reflecting differences in computational resources and application requirements.

Conventional federated fine-tuning relies on a shared model backbone, where clients exchange compatible parameter updates for server-side aggregation (Figure~\ref{fig:overview}a). This requirement limits collaboration among clients using different LLM architectures and model sizes, restricting their ability to retain independently chosen backbones. This raises a central question: \emph{How can clients with different LLM backbones benefit from one another's locally learned knowledge without requiring compatible model parameters?}

To address this challenge, we propose \textbf{Federated Verifier Pool} (\ours{}), a framework that shifts collaboration from parameter aggregation to shared verification. Our key insight is that a verifier can assess the correctness of a candidate response without accessing the parameters of the LLM that generated it \citep{cobbe2021verifiers,lightman2024verify}. By pooling locally trained verifiers rather than aggregating model parameters, \ours{} enables clients with heterogeneous LLM backbones to share learned verification knowledge while retaining their independent model choices.

As illustrated in Figure~\ref{fig:overview}b, each client independently trains a verifier using its local data and candidate responses produced by its frozen LLM backbone, with correctness labels derived from reference answers. The trained verifiers form a shared pool without parameter aggregation, retaining their individual architectures and parameters. During inference, the pool combines verifier judgments to select the best response from multiple candidates generated by a target LLM backbone. The pool can be maintained by a server or formed through direct verifier exchange among clients, enabling flexible collaboration without requiring a shared model architecture.

We evaluate \ours{} across five medical question-answering centers with distinct datasets, following the data organisation of FedCoT \citep{li2025fedcot}. Each center employs a different Qwen or LLaMA backbone spanning 3B--14B parameters, together with a different verifier model. Despite these differences, the resulting pool of five locally trained verifiers can be reused across all five LLM backbones without additional verifier training, outperforming local verification at every center. \ours{} further improves mean accuracy by 1.46--5.79 percentage points over the strongest evaluated federated baselines. These results demonstrate that heterogeneous clients can benefit from shared verification while retaining their independently chosen models. Our main contributions are summarised as follows:

\begin{itemize}
\setlength{\itemsep}{1pt}

\item We propose \ours{}, a new heterogeneous federated learning paradigm that enables clients with different LLM families and sizes to collaborate through locally trained verifiers while retaining their chosen generators.

\item We show that a verifier trained on responses from one client's LLM can assess responses from another's. Clients can therefore exchange verifiers directly, form different pools, and use them locally to select among responses from their own LLMs, making a central aggregation server optional.

\item We validate \ours{} across five medical QA centers with distinct datasets and LLM backbones, showing consistent gains over local verification and superior mean accuracy against federated baselines.
\end{itemize}
% ---------------------------------------------------------------------------
\section{Related Work}
\label{sec:related}

\paragraph{Federated optimization and LLM fine-tuning.}
FedAvg \citep{mcmahan2017fedavg} aggregates locally updated model parameters, while FedProx \citep{li2020federated} and adaptive federated optimizers \citep{reddi2021adaptive} improve optimization under heterogeneous data and client conditions.
For LLMs, FedIT \citep{zhang2023fedit} and FFA-LoRA \citep{sun2024ffalora} aggregate locally trained adapters, while FLoRA \citep{wang2024flora} accommodates heterogeneous LoRA ranks through adapter stacking.
FedPrompt \citep{zhao2022fedprompt} instead uses federated soft-prompt tuning.
These approaches generally require compatible parameters for aggregation, limiting collaboration across different LLM backbones.

\paragraph{Model-heterogeneous federated learning.}
Knowledge distillation enables collaboration across heterogeneous models without parameter aggregation.
FedMD \citep{li2019fedmd} exchanges client predictions on public data and iteratively trains local models toward a shared consensus. FedMKT \citep{fan2025fedmkt} enables mutual knowledge transfer between heterogeneous language models through token alignment and selective bidirectional distillation.
However, both methods rely on shared reference data and iterative model updates, introducing additional data requirements that may limit their applicability in settings where suitable shared data are unavailable.
In contrast, \ours{} enables collaboration across different LLM backbones by sharing locally trained verifiers without requiring shared distillation data, cross-model token alignment, or updates to participating generators.

\paragraph{Verification and inference-time selection.}
Inference-time scaling improves performance by generating and evaluating multiple candidate solutions \citep{snell2024scaling}.
Self-Consistency \citep{wang2022selfconsistency} selects answers through majority voting, while trained verifiers \citep{cobbe2021verifiers}, process reward models \citep{lightman2024verify}, and preference-trained reward models such as Skywork-Reward \citep{liu2024skywork} evaluate candidate responses.
Reward-model ensembles have also been explored to improve selection robustness \citep{coste2024rmensembles}.
These advances motivate our use of verification as a new mechanism for federated collaboration across heterogeneous LLMs.

\paragraph{Federated verification.}
FedCoT \citep{li2025fedcot} enhances federated reasoning through locally trained discriminators that evaluate candidate reasoning paths.
It combines correctness and entailment-based supervision, stacking discriminator LoRA modules and averaging classification heads to construct a global discriminator.
Although FedCoT accommodates heterogeneous LoRA ranks, it still requires compatible discriminator components for parameter aggregation.
In contrast, \ours{} introduces a heterogeneous federated learning paradigm that combines independently trained verifiers without parameter aggregation, enabling collaboration across different generator and verifier backbones through a common scoring interface.
% ---------------------------------------------------------------------------
% ---------------------------------------------------------------------------
\section{Methodology}
\label{sec:method}

\ours{} enables clients with different LLM backbones to collaborate through a shared pool of locally trained verifiers (Figure~\ref{fig:overview}).
Each client independently trains a verifier using its local data and contributes it to a shared pool without parameter aggregation.
At inference, the pool combines verifier judgments to select among candidate responses generated by a target LLM.
Collaboration therefore operates through a common question--response interface, allowing clients to choose both their LLM backbones and verifier architectures independently.

\paragraph{Problem Definition}
Consider $M$ centers (federated clients).
Center $i$ holds a private dataset $\mathcal{D}_i=\{(x,a)\}$ of questions and reference answers and chooses a local generator $G_i$.
Its verifier $V_i(x,r;\theta_i)\in\mathbb{R}$ maps a question $x$ and complete response $r$ to a scalar logit.
The generator supplies candidate responses; the verifier is a separately initialised model trained to assess their correctness.
The generators and verifiers may both differ across clients.

\paragraph{Local verifier training and pooling.}
Each client independently fine-tunes its verifier using LoRA \citep{hu2022lora} with locally generated responses and correctness supervision.
For each question--answer pair $(x,a)\in\mathcal{D}_i$, the local generator $G_i$ samples $N_{\mathrm{tr}}$ candidate responses.
Each candidate is assigned a binary label $y\in\{0,1\}$ according to the correctness of its extracted final answer, with incorrect and unparsable responses assigned negative labels.
All labeled candidates are retained to form the local training set $\mathcal{C}_i$.

The verifier predicts the correctness probability $p_i(x,r)=\sigma(V_i(x,r;\theta_i))$ and is optimized using binary cross-entropy:
\begin{equation}
\mathcal{L}_i =
\frac{1}{|\mathcal{C}_i|}
\sum_{(x,r,y)\in\mathcal{C}_i}
\ell_{\mathrm{BCE}}(p_i(x,r),y).
\label{eq:bce}
\end{equation}
Training updates the verifier's LoRA adapters and scoring components, while the pretrained verifier backbone and local generator remain frozen.

After training, each client contributes its trained verifier to a shared pool $\mathcal{V}=\{V_1,\ldots,V_M\}$. The pool can be maintained by a server or formed through direct verifier sharing among participating clients, without parameter aggregation. Each verifier retains its own backbone and scoring architecture, allowing heterogeneous verifiers to participate in collective verification. Local training data and generators remain at their respective clients.

\paragraph{Collective verification.}
At inference, a target generator $G_m$, such as a participating client's LLM, samples $N$ candidate responses $\{r_1,\ldots,r_N\}$ for a given question $x$, each containing a reasoning trace and a proposed answer.
Every verifier in the shared pool evaluates these candidates and produces a correctness score.
The collective score is computed by averaging the sigmoid-transformed scores of all $M$ verifiers:
\begin{equation}
S(x,r_j)=\frac{1}{M}\sum_{i=1}^{M}
\sigma\bigl(V_i(x,r_j;\theta_i)\bigr).
\label{eq:pool}
\end{equation}
The candidate with the highest collective score is selected:
\begin{equation}
j^\star=\arg\max_{1\leq j\leq N} S(x,r_j).
\end{equation}
The final answer is extracted from the selected response $r_{j^\star}$.

Once constructed, the same verifier pool can evaluate candidate responses from different target generators without additional verifier training.
This enables clients to benefit from collective verification while retaining their chosen generator backbones.
% ---------------------------------------------------------------------------
\section{Experimental Setup}
\label{sec:setup}

\paragraph{Datasets and centers.}
We follow the five-center medical question-answering (QA) setting of FedCoT \citep{li2025fedcot}, treating five public datasets as separate federated clients: BioASQ yes/no \citep{tsatsaronis2015bioasq}, MedMCQA \citep{pal2022medmcqa}, MedQA \citep{jin2020medqa}, medical MMLU \citep{hendrycks2020mmlu}, and PubMedQA \citep{jin2019pubmedqa}.
These datasets vary in question formats, answer types, and distributions, comprising 14{,}421 training and 6{,}243 test questions in total.
Dataset statistics and preprocessing details are provided in Appendix~\ref{app:data}.

\begin{table}[!ht]
\centering\footnotesize
\renewcommand{\arraystretch}{1.12}
\begin{tabular*}{\columnwidth}{@{}l@{\extracolsep{\fill}}ll@{}}
\toprule
Center & Backbone & Verifier \\
\midrule
BioASQ   & LLaMA-3.2-3B & \shortstack[l]{GRM\\(LLaMA-3.1-8B)} \\ \addlinespace[2pt]
MedMCQA  & Qwen2.5-7B   & \shortstack[l]{Skywork-Llama v0.2\\(LLaMA-3.1-8B)} \\ \addlinespace[2pt]
MedQA    & LLaMA-3-8B   & \shortstack[l]{OffsetBias\\(LLaMA-3-8B)} \\ \addlinespace[2pt]
MMLU-Med & Qwen2.5-3B   & \shortstack[l]{ArmoRM\\(LLaMA-3-8B)} \\ \addlinespace[2pt]
PubMedQA & Qwen2.5-14B  & \shortstack[l]{Skywork-Qwen3 V2\\(Qwen3-8B)} \\
\bottomrule
\end{tabular*}
\caption{Backbone and locally trained verifier at each center in the Heterogeneous setting.}
\label{tab:client_models}
\end{table}

\paragraph{Configurations of federated verifier pool.}
Each client trains a verifier on responses generated by its local LLM backbone. At inference, the trained verifier pool evaluates candidate responses from a target backbone, which may differ from those used for verifier training.

In \emph{Heterogeneous}, each center uses the LLM backbone and verifier listed in Table~\ref{tab:client_models}.
The LLM backbones span the Qwen and LLaMA families (3B--14B parameters), while the verifiers differ in pretrained checkpoints and scoring architectures.
The resulting pool of five independently trained verifiers is evaluated across all five participating LLM backbones without retraining.

In \emph{Uniform}, all centers use the same LLM backbone and verifier architecture but independently train their verifiers on local data.
We construct separate five-verifier pools using LLaMA-3-8B-Instruct \citep{llama3instruct8b} and Qwen2.5-7B-Instruct \citep{qwen25instruct7b} for local response generation, with verifiers initialized from Skywork-Llama v0.2 \citep{liu2024skywork,skyworkv02}. This configuration examines whether verifier pooling also benefits clients operating under a shared-backbone setting.

\begin{table*}[!t]
\centering\small
\setlength{\tabcolsep}{3.7pt}
\begin{tabular}{lrrrrrr}
\toprule
Method & BioASQ & MedMCQA & MedQA & MMLU-Med & PubMedQA & Avg. \\
\midrule
\multicolumn{7}{l}{\textbf{Target backbone: LLaMA-3-8B-Instruct}} \\
FedIT & 86.29 & 57.47 & 60.80 & 66.26 & 76.60 & 69.48 \\
FFA-LoRA & 91.13 & 58.59 & 61.59 & 65.64 & 76.40 & 70.67 \\
FLoRA & 87.90 & 57.69 & 61.35 & 63.19 & 75.60 & 69.15 \\
FedPrompt & 92.74 & \textbf{59.77} & 61.74 & 69.33 & 77.60 & 72.24 \\
FedProx & 83.87 & 46.74 & 54.52 & 60.12 & 72.20 & 63.49 \\
FedAdam & 81.45 & 44.32 & 48.55 & 53.99 & 52.80 & 56.22 \\
FedAdagrad & 85.48 & 45.64 & 49.33 & 53.99 & 53.80 & 57.65 \\
FedCoT & 88.71 & 46.40 & 56.09 & 60.12 & 76.80 & 65.62 \\
\rowcolor{oursrow} \ours{} (Uniform) & \textbf{94.35} & 58.12 & \textbf{67.95} & \textbf{70.55} & 78.40 & \textbf{73.87} \\
\rowcolor{oursrow} \ours{} (Heterogeneous) & \textbf{94.35} & 58.24 & 66.77 & \textbf{70.55} & \textbf{78.60} & 73.70 \\
\midrule
\multicolumn{7}{l}{\textbf{Target backbone: Qwen2.5-7B-Instruct}} \\
FedIT & 86.29 & 49.82 & 52.16 & 63.19 & 65.00 & 63.29 \\
FFA-LoRA & 88.71 & 46.81 & 52.47 & 67.48 & 68.40 & 64.78 \\
FLoRA & 89.52 & 52.38 & 56.40 & 67.48 & 70.00 & 67.16 \\
FedPrompt & 87.90 & 46.38 & 53.57 & 60.74 & 63.60 & 62.44 \\
FedProx & 72.58 & 43.58 & 47.45 & 62.58 & 56.80 & 56.60 \\
FedAdam & 90.32 & 48.46 & 56.48 & 70.55 & 69.00 & 66.96 \\
FedAdagrad & 87.90 & 48.86 & 56.95 & 70.55 & 71.20 & 67.09 \\
FedCoT & 93.55 & 51.85 & 58.99 & 74.23 & 71.40 & 70.01 \\
\rowcolor{oursrow} \ours{} (Uniform) & 95.97 & 59.60 & 68.11 & \textbf{76.07} & 77.40 & 75.43 \\
\rowcolor{oursrow} \ours{} (Heterogeneous) & \textbf{96.77} & \textbf{60.27} & \textbf{68.26} & \textbf{76.07} & \textbf{77.60} & \textbf{75.80} \\
\bottomrule
\end{tabular}
\caption{Accuracy (\%) across five medical QA datasets. Each table block reports results from all methods using the same target backbone for inference. Avg.\ is the unweighted mean across datasets.}
\label{tab:main}
\end{table*}

\paragraph{Federated baselines.}
We compare \ours{} against ten federated baselines spanning four approaches.
Classical federated optimization methods include FedProx \citep{li2020federated}, FedAdam, and FedAdagrad \citep{reddi2021adaptive}, implemented with LoRA-based backbone adaptation.
Federated LLM fine-tuning methods include FedIT \citep{zhang2023fedit}, FFA-LoRA \citep{sun2024ffalora}, FLoRA \citep{wang2024flora}, and FedPrompt \citep{zhao2022fedprompt}.
FedCoT \citep{li2025fedcot} serves as a federated discriminator-based reasoning baseline.
These eight methods use the original client-local data partitions without shared public training data.

We additionally include FedMD \citep{li2019fedmd} and FedMKT \citep{fan2025fedmkt} as model-heterogeneous federated learning baselines.
Both rely on shared reference data for knowledge distillation.
For these two methods, we reserve approximately 10\% of the original training questions as a shared labeled reference set, yielding 1,442 shared and 12,979 private questions.
Each client accesses its private partition and the shared reference set, whereas \ours{} does not require shared training questions.
For this comparison, all three methods use the heterogeneous client backbones listed in Table~\ref{tab:client_models}.
FedMD uses answer-option prediction distillation, while FedMKT employs token-aligned bidirectional distillation with an additional server-side LLM.
Further details are provided in Appendix~\ref{app:baselines}.

\paragraph{Implementation and evaluation.}
For \ours{}, each client generates eight training candidates per question and fine-tunes its verifier using rank-16 LoRA \citep{hu2022lora} and trainable scoring components.
Training uses AdamW for two epochs at a learning rate of $10^{-4}$, while the generator and pretrained verifier backbone remain frozen.

For comparisons with federated baselines, we evaluate all methods using LLaMA-3-8B-Instruct and Qwen2.5-7B-Instruct as target backbone architectures.
In \emph{Uniform}, each target backbone is evaluated using a verifier pool trained on responses from that same backbone.
In \emph{Heterogeneous}, the single pool formed from the five clients' locally trained verifiers is reused for both targets without retraining. At inference, \ours{} selects among eight candidates using pooled verifier scores. Further details are provided in Appendix~\ref{app:implementation}.

Following FedCoT \citep{li2025fedcot}, we use accuracy as the primary evaluation metric.
A prediction is correct if its extracted final answer matches the normalized reference answer; invalid or unparsable responses count as incorrect.
We report accuracy on each center's full test set and the unweighted mean across all five centers.

% ---------------------------------------------------------------------------
\section{Results}
\label{sec:results}

We evaluate \ours{} as a federated learning paradigm that enables clients with heterogeneous LLMs to collaborate by sharing locally trained verifiers.

\begin{table*}[!t]
\centering\small
\setlength{\tabcolsep}{3.7pt}
\begin{tabular}{lrrrrrr}
\toprule
Method & BioASQ & MedMCQA & MedQA & MMLU-Med & PubMedQA & Avg. \\
\midrule
\multicolumn{7}{l}{\textbf{Target backbone: LLaMA-3-8B-Instruct}} \\
FedMD ($N=1$) & \textbf{95.97} & 57.06 & 61.90 & 65.64 & 76.20 & 71.36 \\
FedMD + SC ($N=8$) & 93.55 & 57.52 & 61.98 & 66.26 & 75.40 & 70.94 \\
FedMKT ($N=1$) & \textbf{95.97} & 55.49 & 61.59 & 65.64 & 73.80 & 70.50 \\
FedMKT + SC ($N=8$) & \textbf{95.97} & 56.90 & 61.59 & 63.80 & 75.20 & 70.69 \\
\rowcolor{oursrow} \ours{} (Heterogeneous) & 94.35 & \textbf{58.24} & \textbf{66.77} & \textbf{70.55} & \textbf{78.60} & \textbf{73.70} \\
\midrule
\multicolumn{7}{l}{\textbf{Target backbone: Qwen2.5-7B-Instruct}} \\
FedMD ($N=1$) & 85.48 & 50.11 & 60.49 & 66.26 & 70.40 & 66.55 \\
FedMD + SC ($N=8$) & 91.94 & 54.63 & 65.20 & 73.62 & 74.60 & 72.00 \\
FedMKT ($N=1$) & 93.55 & 52.09 & 58.05 & \textbf{76.69} & 70.80 & 70.24 \\
FedMKT + SC ($N=8$) & 94.35 & 53.96 & 57.66 & 76.07 & 73.40 & 71.09 \\
\rowcolor{oursrow} \ours{} (Heterogeneous) & \textbf{96.77} & \textbf{60.27} & \textbf{68.26} & 76.07 & \textbf{77.60} & \textbf{75.80} \\
\bottomrule
\end{tabular}
\caption{Accuracy (\%) across five medical QA datasets. FedMD, FedMKT, and \ours{} use the same five client backbones listed in Table~\ref{tab:client_models}, with the same target backbone architectures for inference as in Table~\ref{tab:main}. FedMD and FedMKT additionally use a shared public reference set for knowledge distillation, whereas \ours{} uses only client-local training data. Avg.\ is the unweighted mean across datasets.}
\label{tab:model_heterogeneous}
\end{table*}

% Table fragment for the manuscript.
% Requires the manuscript's existing booktabs, xcolor (table option), and \ours macro.
\begin{table*}[!t]
\centering
\small
\setlength{\tabcolsep}{4pt}
\renewcommand{\arraystretch}{1.05}

\begin{tabular}{lccccc@{\hspace{14pt}}c}
\toprule
& Client A & Client B & Client C & Client D & Client E & Pool \\
\cmidrule(lr){2-6}\cmidrule(lr){7-7}
Target backbone
& \shortstack{BioASQ\\GRM}
& \shortstack{MedMCQA\\Skywork-Llama}
& \shortstack{MedQA\\OffsetBias}
& \shortstack{MMLU-Med\\ArmoRM}
& \shortstack{PubMedQA\\Skywork-Qwen3}
& \shortstack{\ours{}\\(Heterogeneous)} \\
\midrule

Llama-3.2-3B & \cellcolor[HTML]{FFF8DC}69.69 & 70.95 & 67.51 & 68.18 & 71.52 & \textbf{72.85} \\[-2pt]
& \cellcolor[HTML]{FFF8DC}{\footnotesize(65.74)} & & & & & \\[6pt]

Qwen2.5-7B & 74.08 & \cellcolor[HTML]{FFF8DC}74.90 & 72.66 & 73.49 & 74.94 & \textbf{75.80} \\[-2pt]
& & \cellcolor[HTML]{FFF8DC}{\footnotesize(71.25)} & & & & \\[6pt]

Llama-3-8B & 72.09 & 71.81 & \cellcolor[HTML]{FFF8DC}70.56 & 68.42 & 72.73 & \textbf{73.70} \\[-2pt]
& & & \cellcolor[HTML]{FFF8DC}{\footnotesize(60.82)} & & & \\[6pt]

Qwen2.5-3B & 67.00 & 68.48 & 65.58 & \cellcolor[HTML]{FFF8DC}65.82 & 67.37 & \textbf{70.21} \\[-2pt]
& & & & \cellcolor[HTML]{FFF8DC}{\footnotesize(64.10)} & & \\[6pt]

Qwen2.5-14B & 77.18 & 77.77 & 77.19 & 76.88 & \cellcolor[HTML]{FFF8DC}78.85 & \textbf{78.90} \\[-2pt]
& & & & & \cellcolor[HTML]{FFF8DC}{\footnotesize(74.46)} & \\

\bottomrule
\end{tabular}

\caption{Five-dataset mean accuracy (\%) for cross-backbone verifier evaluation. Shaded diagonal cells correspond to each center's assigned backbone and verifier; parentheses report pretrained verifier results.}
\label{tab:cross_generator}
\end{table*}

\paragraph{Comparison with federated baselines.}
We first compare our proposed paradigm with eight conventional federated baselines across five medical QA datasets, each representing an independent federated center.
Specifically, we evaluate the methods using LLaMA-3-8B-Instruct and Qwen2.5-7B-Instruct as target backbones for inference.
Within our paradigm, \ours{} (Uniform) uses the same generator backbone across centers, whereas \ours{} (Heterogeneous) enables centers using backbones from different model families and at different scales to collaborate through a shared verifier pool, as illustrated in Table~\ref{tab:client_models}.
Both configurations outperform the baselines on most datasets and achieve higher mean accuracy than every baseline for both target backbones.
Notably, with Qwen2.5-7B-Instruct as the target backbone, Heterogeneous and Uniform achieve mean accuracies of 75.80\% and 75.43\%, respectively, surpassing the strongest baseline by 5.79 and 5.42 percentage points.

We further compare \ours{} (Heterogeneous) with FedMD and FedMKT, which support collaboration across heterogeneous client models through knowledge distillation while relying on shared reference data and iterative updates to the participating LLMs.
Their reliance on a shared reference dataset imposes an additional prerequisite for collaboration, limiting their applicability in federated settings where suitable shared data are unavailable. To accommodate this requirement, we conduct controlled experiments by reserving a subset of the training questions as a shared reference set for reproducing these methods and comparing them with \ours{}, as shown in Table~\ref{tab:model_heterogeneous}. Specifically, \ours{} achieves higher mean accuracy on both target backbones than FedMD and FedMKT in both their original single-response configurations and their Self-Consistency (SC) variants.

These results demonstrate that \ours{} provides an effective alternative to conventional federated fine-tuning: clients collaborate by sharing locally learned verification knowledge while retaining their local data and independently chosen model architectures.

\begin{table*}[!t]
\centering\small
\setlength{\tabcolsep}{4pt}
\begin{tabular}{lrrrrrrr}
\toprule
Selector & BioASQ & MedMCQA & MedQA & MMLU-Med & PubMedQA & Avg. & $\Delta$ (pp) \\
\midrule
\multicolumn{8}{l}{\textbf{Target backbone: LLaMA-3-8B-Instruct}} \\
\ours{} (Teacher pool) & 94.35 & \textbf{58.24} & 66.77 & 70.55 & \textbf{78.60} & 73.70 & \textemdash \\
Distilled verifier (Public data) & \textbf{96.77} & 57.85 & 68.34 & 71.17 & \textbf{78.60} & 74.55 & \textcolor{green!45!black}{$+0.84$} \\
Distilled verifier (Local data) & \textbf{96.77} & 57.16 & \textbf{68.66} & \textbf{73.62} & 78.00 & \textbf{74.84} & \textcolor{green!45!black}{$+1.14$} \\
\midrule
\multicolumn{8}{l}{\textbf{Target backbone: Qwen2.5-7B-Instruct}} \\
\ours{} (Teacher pool) & \textbf{96.77} & \textbf{60.27} & \textbf{68.26} & 76.07 & 77.60 & 75.80 & \textemdash \\
Distilled verifier (Public data) & 95.97 & 58.91 & \textbf{68.26} & 77.91 & 77.60 & 75.73 & \textcolor{red!75!black}{$-0.07$} \\
Distilled verifier (Local data) & 95.97 & 59.00 & 67.87 & \textbf{79.14} & \textbf{78.00} & \textbf{76.00} & \textcolor{green!45!black}{$+0.20$} \\
\bottomrule
\end{tabular}
\caption{Distillation of the Heterogeneous verifier pool. Public distillation produces one shared verifier, whereas local distillation produces a separate verifier at each center. Teacher and students select from the same eight candidates per question using mean pool probabilities and single-verifier sigmoid scores, respectively. Values are accuracy (\%); Avg.\ equally weights the five datasets. Best results in each block are bold. $\Delta$ denotes the change in average accuracy relative to the teacher pool (pp), computed before rounding.}
\label{tab:distillation}
\end{table*}

\paragraph{Cross-model federated collaboration.}
We next evaluate cross-model federated collaboration under \ours{} by directly exchanging locally trained verifiers among clients, as reported in Table~\ref{tab:cross_generator}. Specifically, the pale-yellow diagonal cells identify each center's local training configuration: the frozen target backbone used to generate training responses, the selected verifier model, and the local dataset. The values in parentheses report the corresponding verifier's performance before local training.
Each row corresponds to inference with a participating center's target backbone, and their performance if other center's verifier are applied. The final column reports the results obtained with the shared \ours{} verifier pool.

Consider Client A for example, which uses BioASQ for local training, LLaMA-3.2-3B as its frozen generator, and GRM as its verifier. For responses generated by this backbone, mean accuracy increases from 65.74\% with the pretrained GRM to 69.69\% after local verifier training. The same locally trained GRM can then be directly used by Client B to evaluate responses from its Qwen2.5-7B backbone, yielding a mean accuracy of 74.08\%, compared with 71.25\% when using Client B's own pretrained Skywork-Llama verifier. By using the shared \ours{} pool, Client A can further draw on the verification knowledge contributed by all participating clients, reaching a mean accuracy of 72.85\% with its LLaMA-3.2-3B backbone.

These results illustrate how \ours{} realizes federated collaboration through shared verification. Each client learns verification knowledge locally, contributes it through a trained verifier, and can directly use other clients' verifiers individually or collectively through the pool. This exchange allows clients to benefit from one another's local learning while keeping their data local and retaining their independently chosen models, embodying the core idea of federated learning.

\begin{table*}[!t]
\centering\small
\setlength{\tabcolsep}{4pt}
\begin{tabular}{lrrrrrr}
\toprule
Selector & BioASQ & MedMCQA & MedQA & MMLU-Med & PubMedQA & Avg. \\
\midrule
\multicolumn{7}{l}{\textbf{Evaluation generator: LLaMA-3-8B-Instruct}} \\
CoT ($N=1$) & 81.45 & 41.50 & 45.56 & 49.69 & 55.80 & 54.80 \\
Self-Consistency ($N=8$) & 93.55 & 52.21 & 62.06 & 62.58 & 77.40 & 69.56 \\
Pretrained single verifier ($N=8$) & 80.65 & 50.56 & 58.68 & 59.51 & 59.40 & 61.76 \\
Pretrained verifier pool ($N=8$) & 80.65 & 50.97 & 57.66 & 61.96 & 57.00 & 61.65 \\
\rowcolor{oursrow} \ours{} (Uniform) ($N=8$) & 94.35 & 58.12 & 67.95 & 70.55 & 78.40 & 73.87 \\
\rowcolor{oursrow} \ours{} (Heterogeneous) ($N=8$) & 94.35 & 58.24 & 66.77 & 70.55 & 78.60 & 73.70 \\
\midrule
\multicolumn{7}{l}{\textbf{Evaluation generator: Qwen2.5-7B-Instruct}} \\
CoT ($N=1$) & 91.94 & 49.80 & 56.17 & 69.33 & 71.80 & 67.80 \\
Self-Consistency ($N=8$) & 94.35 & 55.46 & 64.57 & 76.07 & 74.80 & 73.05 \\
Pretrained single verifier ($N=8$) & 91.94 & 53.72 & 62.61 & 73.01 & 75.00 & 71.25 \\
Pretrained verifier pool ($N=8$) & 92.74 & 54.98 & 64.57 & 73.62 & 75.60 & 72.30 \\
\rowcolor{oursrow} \ours{} (Uniform) ($N=8$) & 95.97 & 59.60 & 68.11 & 76.07 & 77.40 & 75.43 \\
\rowcolor{oursrow} \ours{} (Heterogeneous) ($N=8$) & 96.77 & 60.27 & 68.26 & 76.07 & 77.60 & 75.80 \\
\bottomrule
\end{tabular}
\caption{Ablation study of answer-selection strategies across five medical QA datasets. Within each table block, all methods use the same evaluation generator and saved candidate responses. CoT uses the first response ($N=1$), while Self-Consistency, the pretrained verifiers, and the trained verifier pools select from the same eight candidates ($N=8$). Verifier-based selectors use sigmoid scores, averaged across members for pools; exact ties select the earliest candidate. Pretrained single verifier refers to Skywork-Llama v0.2 before local fine-tuning; Pretrained verifier pool combines the five released heterogeneous verifiers without local fine-tuning. Avg.\ is the unweighted mean accuracy (\%) across datasets.}
\label{tab:training_full}
\end{table*}

\paragraph{Distillation for efficient inference.}
To reduce the inference cost of the shared pool, we explore distilling its collective verification knowledge into a single verifier.
We investigate this as an optional deployment strategy and report the distilled variants separately from the main \ours{} results.
We use the original Heterogeneous pool as the teacher and initialize each student verifier from pretrained Skywork. We consider two distillation settings: public distillation trains one shared student using a shared reference set, whereas local distillation trains a separate student at each client using its local dataset.
All students are trained for two epochs using binary cross-entropy with the teacher pool's mean probabilities as soft targets, while the teacher pool and generators remain frozen.

As shown in Table~\ref{tab:distillation}, the public student achieves mean accuracies of 74.55\% on LLaMA and 75.73\% on Qwen, compared with 73.70\% and 75.80\% for the teacher pool.
The local students achieve corresponding mean accuracies of 74.84\% and 76.00\%.
Both distillation settings retain performance comparable to the teacher pool.
The small gains may reflect the additional fine-tuning stage introduced by distillation.
These results demonstrate the feasibility of transferring the pool's collective verification knowledge into a single verifier for subsequent inference at each client.

% ---------------------------------------------------------------------------
\section{Ablation study}
\label{sec:analysis}

We examine how client-local verifier training, candidate budget, and score aggregation affect answer selection, and report the communication and inference costs of verifier pooling.

\begin{table}[!b]
\centering\small
\setlength{\tabcolsep}{3pt}
\begin{tabular}{lrrrr}
\toprule
& \multicolumn{2}{c}{\textbf{Uniform}} & \multicolumn{2}{c}{\textbf{Heterogeneous}} \\
\cmidrule(lr){2-3}\cmidrule(lr){4-5}
Aggregation rule & LLaMA & Qwen & LLaMA & Qwen \\
\midrule
Mean probability & 73.87 & \textbf{75.43} & 73.70 & \textbf{75.80} \\
Mean logit & \textbf{74.21} & 75.42 & \textbf{74.25} & 75.77 \\
Max probability & 73.33 & 74.41 & 73.68 & 75.32 \\
Member majority vote & 73.82 & 74.96 & 73.40 & 75.27 \\
\bottomrule
\end{tabular}
\caption{Ablation study of verifier aggregation strategies. Values report five-dataset mean accuracy (\%) under Uniform and Heterogeneous configurations using LLaMA-3-8B-Instruct and Qwen2.5-7B-Instruct as target backbones. Member majority vote selects the candidate receiving the most verifier votes.}
\label{tab:ablation}
\end{table}

\begin{figure}[!t]
\centering
\resizebox{\columnwidth}{!}{%
\begin{tikzpicture}
\begin{groupplot}[
  group style={group size=2 by 1, horizontal sep=4mm, y descriptions at=edge left},
  width=0.56\columnwidth, height=4.6cm,
  xmode=log, log basis x=2, xtick={1,2,4,8}, xticklabels={1,2,4,8},
  xlabel={$N$ candidates}, ymin=52, ymax=86,
  tick label style={font=\scriptsize}, label style={font=\scriptsize}, title style={font=\scriptsize\bfseries, yshift=-1mm},
  grid=major, grid style={gray!20},
  legend style={font=\scriptsize, at={(1.05,-0.33)}, anchor=north, legend columns=2, draw=none, column sep=3pt},
]
\nextgroupplot[title={LLaMA-3-8B-Instruct}, ylabel={Avg.\ accuracy (\%)}]
\addplot[cB, very thick, mark=*, mark size=1.5pt] coordinates {(1,54.8015689280)(2,66.4061760427)(4,71.3338201663)(8,73.8745791129)};
\addplot[violet, very thick, dashed, mark=diamond*, mark size=1.8pt] coordinates {(1,54.8015689280)(2,66.3128878709)(4,70.9307195884)(8,73.7028216139)};
\addplot[cC, thick, mark=square*, mark size=1.3pt] coordinates {(1,54.8015689280)(2,63.3130286891)(4,67.9111911837)(8,69.5589072388)};
\addplot[cA, thick, mark=triangle*, mark size=1.6pt] coordinates {(1,54.8015689280)(2,57.7799464586)(4,60.6936629402)(8,61.7592888587)};
\addplot[cD, dashed, mark=o, mark size=1.3pt] coordinates {(1,54.8015689280)(2,69.5001576028)(4,77.5334182254)(8,83.8936796207)};
\legend{\ours{} (Uniform), \ours{} (Heterogeneous), Self-Consistency, Pretrained Verifier, Oracle}
\nextgroupplot[title={Qwen2.5-7B-Instruct}]
\addplot[cB, very thick, mark=*, mark size=1.5pt] coordinates {(1,67.8047939090)(2,71.7503834627)(4,74.3308640691)(8,75.4293140379)};
\addplot[violet, very thick, dashed, mark=diamond*, mark size=1.8pt] coordinates {(1,67.8047939090)(2,71.9701883101)(4,74.9780600090)(8,75.7959014079)};
\addplot[cC, thick, mark=square*, mark size=1.3pt] coordinates {(1,67.8047939090)(2,70.0845616377)(4,71.4425746698)(8,73.0525844913)};
\addplot[cA, thick, mark=triangle*, mark size=1.6pt] coordinates {(1,67.8047939090)(2,70.2079343326)(4,71.7751156470)(8,71.2534118185)};
\addplot[cD, dashed, mark=o, mark size=1.3pt] coordinates {(1,67.8047939090)(2,74.9366241084)(4,80.9944413641)(8,84.8017911488)};
\end{groupplot}
\end{tikzpicture}}
\caption{Candidate-budget scaling analysis across two target backbones. All selection methods use the same first $N\in\{1,2,4,8\}$ saved candidates, with model parameters fixed throughout. SC uses majority voting, and Pretrained Verifier denotes Skywork-Llama without local fine-tuning and ranks candidates by sigmoid score, with exact ties resolved by candidate order. At $N=1$, all methods coincide with CoT. Oracle indicates whether at least one of the first $N$ candidates is correct.}
\label{fig:scaling}
\end{figure}
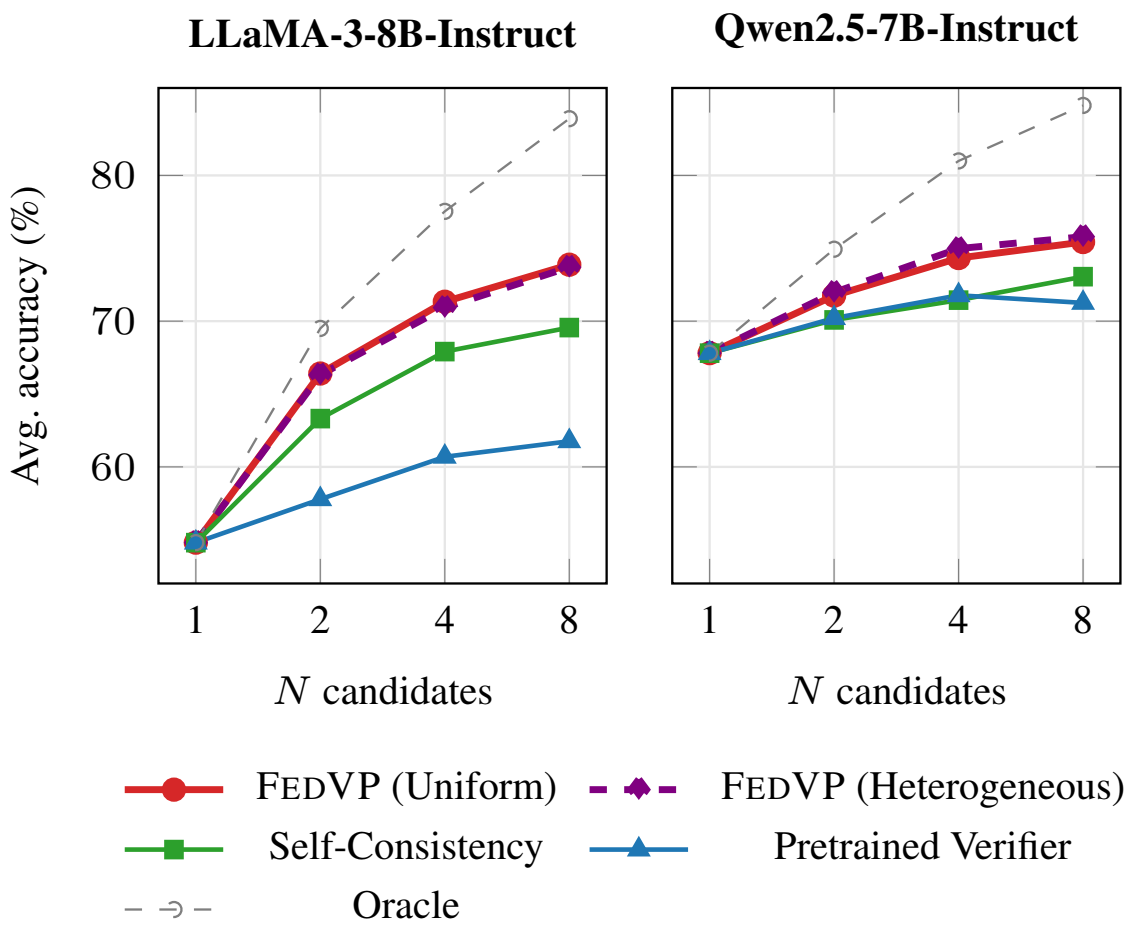

\paragraph{Effectiveness of trained verifier pools.}
We first examine the contribution of client-local verifier training through the progressive ablation in Table~\ref{tab:training_full}.
The comparison includes single-response CoT, SC with majority voting over eight responses, a pretrained single verifier, and a pretrained heterogeneous pool, with both \ours{} configurations serving as locally trained references.
SC raises mean accuracy from 54.80\% to 69.56\% on LLaMA and from 67.80\% to 73.05\% on Qwen, illustrating the potential of inference-time scaling through multiple candidate responses.
On the same eight candidates, the pretrained single verifier achieves 61.76\% and 71.25\%, while the pretrained heterogeneous pool achieves 61.65\% and 72.30\%, respectively.
Both pretrained strategies perform below SC, and the pool remains close to the single verifier, indicating that combining pretrained verifier judgments alone provides limited benefit.
After local training, \ours{} (Heterogeneous) reaches 73.70\% and 75.80\%, and both trained configurations outperform SC.
This improvement highlights the role of client-local training in developing useful verification knowledge, which can then be shared and combined across clients to improve answer selection.

\paragraph{Scaling the candidate budget.}
We next explore how scaling the candidate budget affects inference cost and answer accuracy, as shown in Figure~\ref{fig:scaling}.
We vary $N\in\{1,2,4,8\}$ while keeping all model parameters fixed and using the same first $N$ candidates for all selectors.
At $N=1$, all methods coincide with CoT because only one response is available.
As $N$ increases, both \ours{} configurations improve consistently and outperform SC and pretrained verification at every multi-candidate budget, demonstrating that shared verification can translate additional candidate generation into better answer selection. The more gradual gains at larger budgets illustrate the trade-off between inference cost and accuracy. The oracle curve reports the accuracy achievable by selecting a correct response whenever one exists among the candidates. Its continued rise with $N$ indicates that larger candidate sets provide more opportunities for correct selection.
The remaining gap between \ours{} and the oracle highlights further room for improving this federated learning paradigm through more effective collective verification.

\paragraph{Aggregation of verifier judgments.}
We next compare four aggregation strategies under both Uniform and Heterogeneous configurations: mean probability, mean logit, maximum probability, and member majority voting (Table~\ref{tab:ablation}).
All four strategies yield broadly comparable mean accuracies on both target backbones, indicating that shared verification can be effectively implemented with different simple aggregation rules.
The two averaging strategies perform slightly better overall, with mean logit achieving the highest accuracy on LLaMA and mean probability on Qwen.
We adopt mean-probability aggregation as a simple default.

\begin{table}[!t]
\centering\small
\setlength{\tabcolsep}{3pt}
\begin{tabular}{lrrrr}
\toprule
& \multicolumn{2}{c}{LLaMA-3-8B}
& \multicolumn{2}{c}{Qwen2.5-7B} \\
\cmidrule(lr){2-3}\cmidrule(lr){4-5}
Pool configuration & Avg. & $\Delta$ (pp) & Avg. & $\Delta$ (pp) \\
\midrule
Full pool    & 73.70 & -- & 75.80 & -- \\
w/o Client A & 73.63 & \textcolor{red!75!black}{-0.07}
             & 75.71 & \textcolor{red!75!black}{-0.09} \\
w/o Client B & 73.32 & \textcolor{red!75!black}{-0.38}
             & 75.43 & \textcolor{red!75!black}{-0.37} \\
w/o Client C & 73.54 & \textcolor{red!75!black}{-0.16}
             & 76.00 & \textcolor{green!45!black}{+0.20} \\
w/o Client D & 73.65 & \textcolor{red!75!black}{-0.05}
             & 75.37 & \textcolor{red!75!black}{-0.43} \\
w/o Client E & 72.62 & \textcolor{red!75!black}{-1.08}
             & 75.40 & \textcolor{red!75!black}{-0.39} \\
\bottomrule
\end{tabular}
\caption{Leave-one-out ablation of the Heterogeneous verifier pool, removing each client's contributed verifier. Avg. is the five-dataset mean accuracy (\%); $\Delta$ is the change from the full pool (pp), with decreases in red and increases in green.}
\label{tab:heterogeneous_loo}
\end{table}
\paragraph{Leave-One-Out Verifier Ablation}
We conduct a leave-one-out ablation to examine the contribution of each locally trained verifier to the heterogeneous pool, removing one verifier at a time while retaining the other four. For LLaMA-3-8B, each removal reduces the average accuracy by 0.05--1.08 percentage points. For Qwen2.5-7B, accuracy decreases in four of the five leave-one-out configurations, whereas excluding Client C's verifier (OffsetBias) yields a modest improvement of 0.20 percentage points. Removing Client E's verifier (Skywork-Qwen3) produces the largest accuracy drop on LLaMA-3-8B. Overall, the relatively small performance changes suggest that the pool does not critically depend on any single verifier, although the effect of removing individual members varies.

\paragraph{Cost of verifier pooling.}
In the \emph{Heterogeneous} setting, each center fine-tunes 41.9--48.3 million parameters in its verifier's LoRA adapters and scoring components.
With the pretrained verifier backbones already available, each center shares these trained weights once, corresponding to a transfer of 168--193\,MB in FP32.
At inference, the original pool scores eight candidate responses using all five verifiers.
Deploying a distilled student reduces this to one verifier evaluation per candidate.

% ---------------------------------------------------------------------------
\section{Conclusion}
We introduced \textbf{Federated Verifier Pool} (\ours{}), a new heterogeneous federated learning paradigm that shifts collaboration from parameter aggregation to shared verification.
Through locally trained verifiers, clients with different LLM families and sizes can collaborate while retaining their chosen backbones. Clients can use these verifiers individually or combine them into pools, keeping training data local and generators frozen while making a central aggregation server optional.
Experiments across five medical QA centers demonstrate effective cross-model collaboration and higher mean accuracy than the evaluated conventional and model-heterogeneous federated baselines.

\section*{Limitations}
This work focuses on demonstrating the feasibility of shared verification as an alternative federated learning paradigm. Thus, our evaluation only cover five medical question-answering datasets and LLaMA and Qwen backbones ranging from 3B to 14B parameters. Further evaluation across additional domains and a wider range of model families and sizes would help assess the broader applicability of \ours{}.

% ---------------------------------------------------------------------------
\bibliography{references}

@misc{li2025fedcot,
  title = {{FedCoT}: Communication-Efficient Federated Reasoning Enhancement for Large Language Models},
  author = {Chuan Li and Qianyi Zhao and Fengran Mo and Cen Chen},
  year = {2025},
  eprint = {2508.10020},
  archivePrefix = {arXiv},
  primaryClass = {cs.CL},
  note = {Version 2, revised 6 October 2026},
  url = {https://arxiv.org/abs/2508.10020v2}
}

@article{li2019fedmd,
  title={Fedmd: Heterogenous federated learning via model distillation},
  author={Li, Daliang and Wang, Junpu},
  journal={arXiv preprint arXiv:1910.03581},
  year={2019}
}

@inproceedings{fan2025fedmkt,
  title={Fedmkt: Federated mutual knowledge transfer for large and small language models},
  author={Fan, Tao and Ma, Guoqiang and Kang, Yan and Gu, Hanlin and Song, Yuanfeng and Fan, Lixin and Chen, Kai and Yang, Qiang},
  booktitle={Proceedings of the 31st International Conference on Computational Linguistics},
  pages={243--255},
  year={2025}
}

@misc{wang2022selfconsistency,
  title = {Self-Consistency Improves Chain of Thought Reasoning in Language Models},
  author = {Xuezhi Wang and Jason Wei and Dale Schuurmans and Quoc Le and Ed Chi and Sharan Narang and Aakanksha Chowdhery and Denny Zhou},
  year = {2022},
  eprint = {2203.11171},
  archivePrefix = {arXiv},
  note = {ICLR 2023 camera-ready version available as arXiv v4},
  url = {https://arxiv.org/abs/2203.11171}
}

@misc{zhang2023fedit,
  title = {Towards Building the Federated {GPT}: Federated Instruction Tuning},
  author = {Jianyi Zhang and Saeed Vahidian and Martin Kuo and Chunyuan Li and Ruiyi Zhang and Tong Yu and Yufan Zhou and Guoyin Wang and Yiran Chen},
  year = {2023},
  eprint = {2305.05644},
  archivePrefix = {arXiv},
  note = {Version 2},
  url = {https://arxiv.org/abs/2305.05644v2}
}

@misc{sun2024ffalora,
  title = {Improving {LoRA} in Privacy-preserving Federated Learning},
  author = {Youbang Sun and Zitao Li and Yaliang Li and Bolin Ding},
  year = {2024},
  eprint = {2403.12313},
  archivePrefix = {arXiv},
  note = {Published at ICLR 2024},
  url = {https://arxiv.org/abs/2403.12313}
}

@misc{wang2024flora,
  title = {{FLoRA}: Federated Fine-Tuning Large Language Models with Heterogeneous Low-Rank Adaptations},
  author = {Ziyao Wang and Zheyu Shen and Yexiao He and Guoheng Sun and Hongyi Wang and Lingjuan Lyu and Ang Li},
  year = {2024},
  eprint = {2409.05976},
  archivePrefix = {arXiv},
  note = {Version 1},
  url = {https://arxiv.org/abs/2409.05976v1}
}

@misc{zhao2022fedprompt,
  title = {{FedPrompt}: Communication-Efficient and Privacy Preserving Prompt Tuning in Federated Learning},
  author = {Haodong Zhao and Wei Du and Fangqi Li and Peixuan Li and Gongshen Liu},
  year = {2022},
  eprint = {2208.12268},
  archivePrefix = {arXiv},
  note = {Version 3; the package also retains the ICASSP 2023 accepted PDF},
  url = {https://arxiv.org/abs/2208.12268v3}
}

@misc{liu2024skywork,
  title = {{Skywork-Reward}: Bag of Tricks for Reward Modeling in {LLMs}},
  author = {Chris Yuhao Liu and Liang Zeng and Jiacai Liu and Rui Yan and Jujie He and Chaojie Wang and Shuicheng Yan and Yang Liu and Yahui Zhou},
  year = {2024},
  eprint = {2410.18451},
  archivePrefix = {arXiv},
  url = {https://arxiv.org/abs/2410.18451v1}
}

@article{tsatsaronis2015bioasq,
  title = {An overview of the {BIOASQ} large-scale biomedical semantic indexing and question answering competition},
  author = {George Tsatsaronis and Georgios Balikas and Prodromos Malakasiotis and Ioannis Partalas and Matthias Zschunke and Michael R. Alvers and Dirk Weissenborn and Anastasia Krithara and Sergios Petridis and Dimitris Polychronopoulos and Yannis Almirantis and John Pavlopoulos and Nicolas Baskiotis and Patrick Gallinari and Thierry Arti{\'e}res and Axel-Cyrille Ngonga Ngomo and Norman Heino and Eric Gaussier and Liliana Barrio-Alvers and Michael Schroeder and Ion Androutsopoulos and Georgios Paliouras},
  journal = {BMC Bioinformatics},
  year = {2015},
  volume = {16},
  pages = {138},
  doi = {10.1186/s12859-015-0564-6},
  url = {https://link.springer.com/article/10.1186/s12859-015-0564-6}
}

@misc{pal2022medmcqa,
  title = {{MedMCQA}: A Large-scale Multi-Subject Multi-Choice Dataset for Medical domain Question Answering},
  author = {Ankit Pal and Logesh Kumar Umapathi and Malaikannan Sankarasubbu},
  year = {2022},
  eprint = {2203.14371},
  archivePrefix = {arXiv},
  note = {CHIL 2022},
  url = {https://arxiv.org/abs/2203.14371v1}
}

@misc{jin2020medqa,
  title = {What Disease does this Patient Have? A Large-scale Open Domain Question Answering Dataset from Medical Exams},
  author = {Di Jin and Eileen Pan and Nassim Oufattole and Wei-Hung Weng and Hanyi Fang and Peter Szolovits},
  year = {2020},
  eprint = {2009.13081},
  archivePrefix = {arXiv},
  url = {https://arxiv.org/abs/2009.13081v1}
}

@misc{hendrycks2020mmlu,
  title = {Measuring Massive Multitask Language Understanding},
  author = {Dan Hendrycks and Collin Burns and Steven Basart and Andy Zou and Mantas Mazeika and Dawn Song and Jacob Steinhardt},
  year = {2020},
  eprint = {2009.03300},
  archivePrefix = {arXiv},
  note = {Version 3; ICLR 2021},
  url = {https://arxiv.org/abs/2009.03300v3}
}

@misc{jin2019pubmedqa,
  title = {{PubMedQA}: A Dataset for Biomedical Research Question Answering},
  author = {Qiao Jin and Bhuwan Dhingra and Zhengping Liu and William W. Cohen and Xinghua Lu},
  year = {2019},
  eprint = {1909.06146},
  archivePrefix = {arXiv},
  note = {EMNLP 2019},
  url = {https://arxiv.org/abs/1909.06146v1}
}

@misc{skyworkv02,
  title = {{Skywork-Reward-Llama-3.1-8B-v0.2}: Model Card},
  author = {{Skywork}},
  howpublished = {Hugging Face model repository},
  url = {https://huggingface.co/Skywork/Skywork-Reward-Llama-3.1-8B-v0.2},
  note = {Accessed 23 October 2026; exact checkpoint used in this experiment},
  year = {n.d.}
}

@misc{llama3instruct8b,
  title = {{Meta-Llama-3-8B-Instruct}: Model Card},
  author = {{Meta}},
  howpublished = {Hugging Face model repository},
  url = {https://huggingface.co/meta-llama/Meta-Llama-3-8B-Instruct},
  note = {Accessed 23 October 2026},
  year = {n.d.}
}

@misc{qwen25instruct7b,
  title = {{Qwen2.5-7B-Instruct}: Model Card},
  author = {{Qwen}},
  howpublished = {Hugging Face model repository},
  url = {https://huggingface.co/Qwen/Qwen2.5-7B-Instruct},
  note = {Accessed 23 October 2026},
  year = {n.d.}
}

@inproceedings{mcmahan2017fedavg,
  title = {Communication-Efficient Learning of Deep Networks from Decentralized Data},
  author = {Brendan McMahan and Eider Moore and Daniel Ramage and Seth Hampson and Blaise Ag{\"u}era y Arcas},
  booktitle = {Proceedings of the 20th International Conference on Artificial Intelligence and Statistics},
  series = {Proceedings of Machine Learning Research},
  volume = {54},
  pages = {1273--1282},
  year = {2017}
}

@article{kairouz2021advances,
  title = {Advances and Open Problems in Federated Learning},
  author = {Peter Kairouz and H. Brendan McMahan and others},
  journal = {Foundations and Trends in Machine Learning},
  volume = {14},
  number = {1--2},
  pages = {1--210},
  year = {2021}
}

@inproceedings{hu2022lora,
  title = {{LoRA}: Low-Rank Adaptation of Large Language Models},
  author = {Edward J. Hu and Yelong Shen and Phillip Wallis and Zeyuan Allen-Zhu and Yuanzhi Li and Shean Wang and Lu Wang and Weizhu Chen},
  booktitle = {International Conference on Learning Representations},
  year = {2022}
}

@misc{cobbe2021verifiers,
  title = {Training Verifiers to Solve Math Word Problems},
  author = {Karl Cobbe and Vineet Kosaraju and Mohammad Bavarian and Mark Chen and Heewoo Jun and Lukasz Kaiser and Matthias Plappert and Jerry Tworek and Jacob Hilton and Reiichiro Nakano and Christopher Hesse and John Schulman},
  year = {2021},
  eprint = {2110.14168},
  archivePrefix = {arXiv},
  url = {https://arxiv.org/abs/2110.14168}
}

@inproceedings{lightman2024verify,
  title = {Let's Verify Step by Step},
  author = {Hunter Lightman and Vineet Kosaraju and Yura Burda and Harri Edwards and Bowen Baker and Teddy Lee and Jan Leike and John Schulman and Ilya Sutskever and Karl Cobbe},
  booktitle = {International Conference on Learning Representations},
  year = {2024}
}

@inproceedings{coste2024rmensembles,
  title = {Reward Model Ensembles Help Mitigate Overoptimization},
  author = {Thomas Coste and Usman Anwar and Robert Kirk and David Krueger},
  booktitle = {International Conference on Learning Representations},
  year = {2024}
}

@misc{snell2024scaling,
  title = {Scaling {LLM} Test-Time Compute Optimally can be More Effective than Scaling Model Parameters},
  author = {Charlie Snell and Jaehoon Lee and Kelvin Xu and Aviral Kumar},
  year = {2024},
  eprint = {2408.03314},
  archivePrefix = {arXiv},
  url = {https://arxiv.org/abs/2408.03314}
}

@inproceedings{li2020federated,
  title = {Federated Optimization in Heterogeneous Networks},
  author = {Tian Li and Anit Kumar Sahu and Manzil Zaheer and Maziar Sanjabi and Ameet Talwalkar and Virginia Smith},
  booktitle = {Proceedings of Machine Learning and Systems},
  volume = {2},
  year = {2020},
  url = {https://proceedings.mlsys.org/paper_files/paper/2020/hash/1f5fe83998a09396ebe6477d9475ba0c-Abstract.html}
}

@inproceedings{reddi2021adaptive,
  title = {Adaptive Federated Optimization},
  author = {Sashank Reddi and Zachary Charles and Manzil Zaheer and Zachary Garrett and Keith Rush and Jakub Kone{\v{c}}n{\'y} and Sanjiv Kumar and H. Brendan McMahan},
  booktitle = {International Conference on Learning Representations},
  year = {2021},
  url = {https://openreview.net/forum?id=LkFG3lB13U5}
}
\newpage
\appendix

\section{Data and Evaluation Details}
\label{app:data}

Table~\ref{tab:data} reports the dataset splits and answer formats for the five centers.
BioASQ uses yes/no questions from the Task 7B--11B GoldenEnriched files, with an 80/20 training/test split.
Following the FedCoT preprocessing protocol, MedMCQA uses the first 2{,}000 explanation-bearing questions from the shuffled official training split.
MedQA retains the original splits of the English four-option \texttt{bigbio/med\_qa} dataset.
MMLU-Med combines questions from seven subjects---clinical knowledge, college biology, college medicine, high-school biology, medical genetics, professional medicine, and virology---with an 80/10/10 training/validation/test split.
PubMedQA uses the expert-labeled PQA-L subset.

\begin{table}[!ht]
\centering
\small
\setlength{\tabcolsep}{4pt}
\begin{tabular}{lrrrl}
\toprule
Center & Train & Val. & Test & Format \\
\midrule
BioASQ & 494 & 0 & 124 & yes/no \\
MedMCQA & 2{,}000 & -- & 4{,}183 & 4 options \\
MedQA & 10{,}178 & 1{,}272 & 1{,}273 & 4 options \\
MMLU-Med & 1{,}299 & 162 & 163 & 4 options \\
PubMedQA & 450 & 50 & 500 & yes/no/maybe \\
\midrule
Total & 14{,}421 & & 6{,}243 & \\
\bottomrule
\end{tabular}
\caption{Dataset splits and answer formats. MedMCQA uses its official validation split for testing.}
\label{tab:data}
\end{table}

\subsection{Evaluation Protocol}
\label{app:parser}

\paragraph{Generation and comparison settings.}
Candidates use the FedCoT chain-of-thought prompt.
Temperature and top-$p$ are sampled from $[0.3,0.9]$ and $[0.7,0.95]$ per block of 1{,}024 generations.
Responses reaching the initial limit of 4{,}096 new tokens are regenerated with larger budgets, up to the model's native context limit.
Selectors evaluated with the same fixed generator use identical saved candidates.

\paragraph{Selection and answer extraction.}
CoT returns the first candidate; SC votes over parseable final answers, with ties resolved by first occurrence.
The pretrained single-verifier reference uses Skywork-Reward-Llama-3.1-8B-v0.2 \citep{skyworkv02}.
Pool-score ties select the earliest eligible candidate; member-vote ties use summed member probabilities, then candidate order.
A common reference-independent parser extracts and normalizes explicit final answers for both verifier labels and evaluation, excluding intermediate reasoning.

\section{Implementation Details}
\label{app:implementation}

\paragraph{Additional verifier settings.}
LoRA uses $\alpha=32$ and dropout 0.05 on all attention projections and MLP gate, up, and down projections.
Native scoring components are trained; for ArmoRM, these are the regression and gating layers, with the released reward transformation fixed.
The effective batch size is 32, with a constant schedule, zero weight decay, and no gradient clipping.
Frozen backbones use BF16 and trainable parameters use FP32.
Inputs use official chat templates without truncation.

\subsection{Federated Baselines}
\label{app:baselines}

\paragraph{Generator-tuning baselines.}

We train the generator-tuning baselines for two communication rounds, with all five clients participating in each round.
Each client performs one local epoch per round with an effective batch size of 2. Client contributions are weighted by local training-set size according to each method's aggregation rule. Backbones are frozen in BF16 and trainable parameters use FP32.
Training uses FedCoT prompts, native chat templates, and available explanations followed by canonical answers, without truncation.
FFA-LoRA and FedPrompt compute loss on assistant targets; the other five include both prompt and target.
Table~\ref{tab:baselines} gives method-specific settings.
FedPrompt initializes its soft prompts randomly.

\begin{table}[!t]
\centering
\scriptsize
\setlength{\tabcolsep}{2.5pt}
\begin{tabular}{lcccc}
\toprule
 & FedIT & FFA-LoRA & FLoRA & FedPrompt \\
\midrule
Trainable & LoRA $A,B$ & LoRA $B$ & new $A,B$/round & 20 soft tokens \\
Rank / $\alpha$ & 32 / 16 & 32 / 16 & 32 / 32 & -- \\
Targets & q & q, v & q, v & input emb. \\
Optimizer & AdamW & SGD & AdamW & Adam \\
LR & 3e-4 & 5e-3 & 3e-4 & 0.3 \\
Schedule & linear & const. & linear & const. \\
Grad. clip & 1.0 & -- & 1.0 & -- \\
Aggregation & avg $A$, $B$ & avg $B$ & stack & avg prompt \\
\bottomrule
\end{tabular}
\caption{Method-specific generator-tuning settings. FedProx, FedAdam, and FedAdagrad share FedIT's adapter settings, local optimizer, and initial learning rate; additional settings are given below.}
\label{tab:baselines}
\end{table}

FedProx, FedAdam, and FedAdagrad use the FedIT LoRA configuration with dropout 0.05, zero weight decay, and gradient clipping at 1.0.
FedProx uses $\mu=0.01$.
FedAdam and FedAdagrad use linearly decaying local learning rates and sample-weighted client deltas, with server learning rates $10^{-4}$ and $10^{-3}$, respectively, and $\tau=10^{-3}$.
FedAdam uses $(\beta_1,\beta_2)=(0.9,0.99)$.
Second moments start at $\tau^2$, the FedAdam first moment starts at zero, and bias correction is disabled \citep{reddi2021adaptive}.

\paragraph{FedCoT.}
Longformer-base-4096 discriminators are trained for three rounds of one epoch each \citep{li2025fedcot}.
Client LoRA ranks are $(4,32,32,16,4)$ in Table~\ref{tab:data} order, with $\alpha=r$, dropout 0.1, and query/value targets.
Training uses learning rate $2\times10^{-4}$, effective batch size 16, cosine decay, 10\% warm-up, and weight decay 0.01.
The server stacks adapters and averages classification heads by retained-candidate counts; evaluation uses the final round.

\paragraph{Shared-reference baselines.}
FedMD and FedMKT share 1{,}442 training questions, leaving 12{,}979 private questions.
Each client accesses its private partition and the shared labeled set.
Both methods use independently initialized rank-16 LoRA ($\alpha=32$, dropout 0.05), AdamW at $10^{-4}$, effective batch size 32, and microbatch size 1.

FedMD \citep{li2019fedmd} operates on valid-answer-option logits.
After one public and one private supervised epoch, each of two rounds averages the five clients' raw public-data logits equally, followed by one public epoch minimizing mean absolute error to the consensus and one private supervised epoch.
Supervised targets retain available explanations and canonical answers.

FedMKT \citep{fan2025fedmkt} uses the same split and five client backbones, plus an independently initialized Qwen2.5-7B-Instruct server with public-data access only.
Each of five rounds comprises private supervised client training, public-data distillation to the server, and public-data distillation back to clients, with one epoch per stage.
It exchanges top-128 token logits using MinED alignment and DualMinCE teacher selection, with temperature 1 and loss $0.9\mathcal{L}_{\mathrm{LM}}+0.1\mathcal{L}_{\mathrm{distill}}$.
Both methods are evaluated using the final LLaMA-3-8B and Qwen2.5-7B client models.
\end{document}